\documentclass[sigconf]{acmart}

\usepackage{latexsym}
\usepackage{amsfonts}
\usepackage{algorithm}
\usepackage{algpseudocode}
\usepackage{amsmath}
\usepackage{makecell}
\usepackage{multirow}
\usepackage[T1]{fontenc}
\usepackage[utf8]{inputenc}
\usepackage{microtype}
\usepackage{graphicx}
\usepackage{enumitem}
\AtBeginDocument{%
  }

\begin{document}

\title{Multi-User Chat Assistant (MUCA): a Framework Using LLMs to Facilitate Group Conversations}

\author{Manqing Mao$^*$ Paishun Ting$^*$, Yijian Xiang$^*$\\
Mingyang Xu$^*$, Julia Chen$^*$, Jianzhe Lin$^*$}
\authornote{These authors contributed equally to this research.}
\affiliation{%
  \institution{Microsoft Research}
}
\email{{manqing.mao, paishun.ting, yijianxiang, mingyangxu, juliachen, jianzhelin}@microsoft.com}








\renewcommand{\shortauthors}{}

\vspace{1.2em}
\begin{abstract}
Recent advancements in large language models (LLMs) have provided a new avenue for chatbot development. Most existing research, however, has primarily centered on single-user chatbots that determine \emph{"What"} to answer. This paper highlights the complexity of multi-user chatbots, introducing the \emph{3W} design dimensions: \emph{"What"} to say, \emph{"When"} to respond, and \emph{"Who"} to answer. Additionally, we proposed Multi-User Chat Assistant (MUCA), an LLM-based framework tailored for group discussions. MUCA consists of three main modules: Sub-topic Generator, Dialog Analyzer, and Conversational Strategies Arbitrator. These modules jointly determine suitable response contents, timings, and appropriate addressees. This paper further proposes an LLM-based Multi-User Simulator (MUS) to ease MUCA's optimization, enabling faster simulation of conversations between the chatbot and simulated users, and speeding up MUCA's early development. In goal-oriented conversations with a small to medium number of participants, MUCA demonstrates effectiveness in tasks like chiming in at appropriate timings, generating relevant content, and improving user engagement, as shown by case studies and user studies.
\end{abstract}






\maketitle

\section{Introduction}
\label{sec:introduction}
Recent years have seen a surge of interest in the field of chatbot research. Large language models (LLMs) like GPTs \cite{openai2023gpt4, brown2020language, Radford2019LanguageMA} have emerged as a powerful tool for chatbot development \cite{su2021multi, he2022space}. However, unlike single-user conversation chatbots, limited research on group conversation chatbots restricts their application in tasks like brainstorming sessions and debates.

\begin{figure}[ht]
\vspace{1.2em}
\centering
\includegraphics[width=0.47\textwidth]{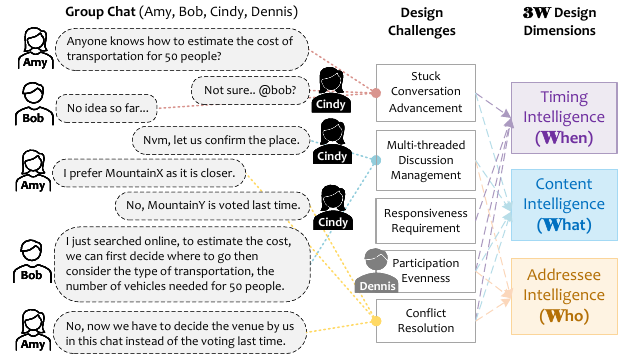}
\vspace{-1.2em}
\caption{A diagram mapping out a group chat sample to its associated five design challenges and further formulated to the proposed \emph{3W} design dimensions.}
\Description{A group chat history that indicates design challenges of a chatbot}
\label{fig:challenges}
\end{figure}

This paper presents \emph{Multi-User Chat Assistant (MUCA)}, an LLM-based framework for group conversation chatbots which, as far as the authors are aware of, is the first LLM-based framework dedicated to multi-user conversations. Unlike single-user chatbots that simply determine \emph{"What"} to answer following a user's inputs, multi-user chatbots have \emph{3W} design dimensions, where the extra two are \emph{"When"} to answer and \emph{"Who"} to answer. We demonstrate that many of the challenges like advancing stuck conversation and managing multi-threaded discussion can be mapped to these \emph{3W} dimensions. To enable fast iteration and development of MUCA, we also devise an LLM-based \emph{Multi-User Simulator (MUS)} that improves over time by leveraging human-in-the-loop feedback.

While MUCA can participate in conversations of chit-chat nature, we demonstrated MUCA’s effectiveness with both case and user studies, focusing on several goal-oriented topics. The evaluation is using quantitative metrics like user engagement, conversation evenness, and opinion consensus. We also measured MUCA's performance by metrics like efficiency, conciseness, and usefulness collected from user feedback, showing that MUCA is superior to a baseline chatbot. The highlights of our work are as follows:

\begin{figure*}[!ht]
\centering
\includegraphics[width=0.96\textwidth]{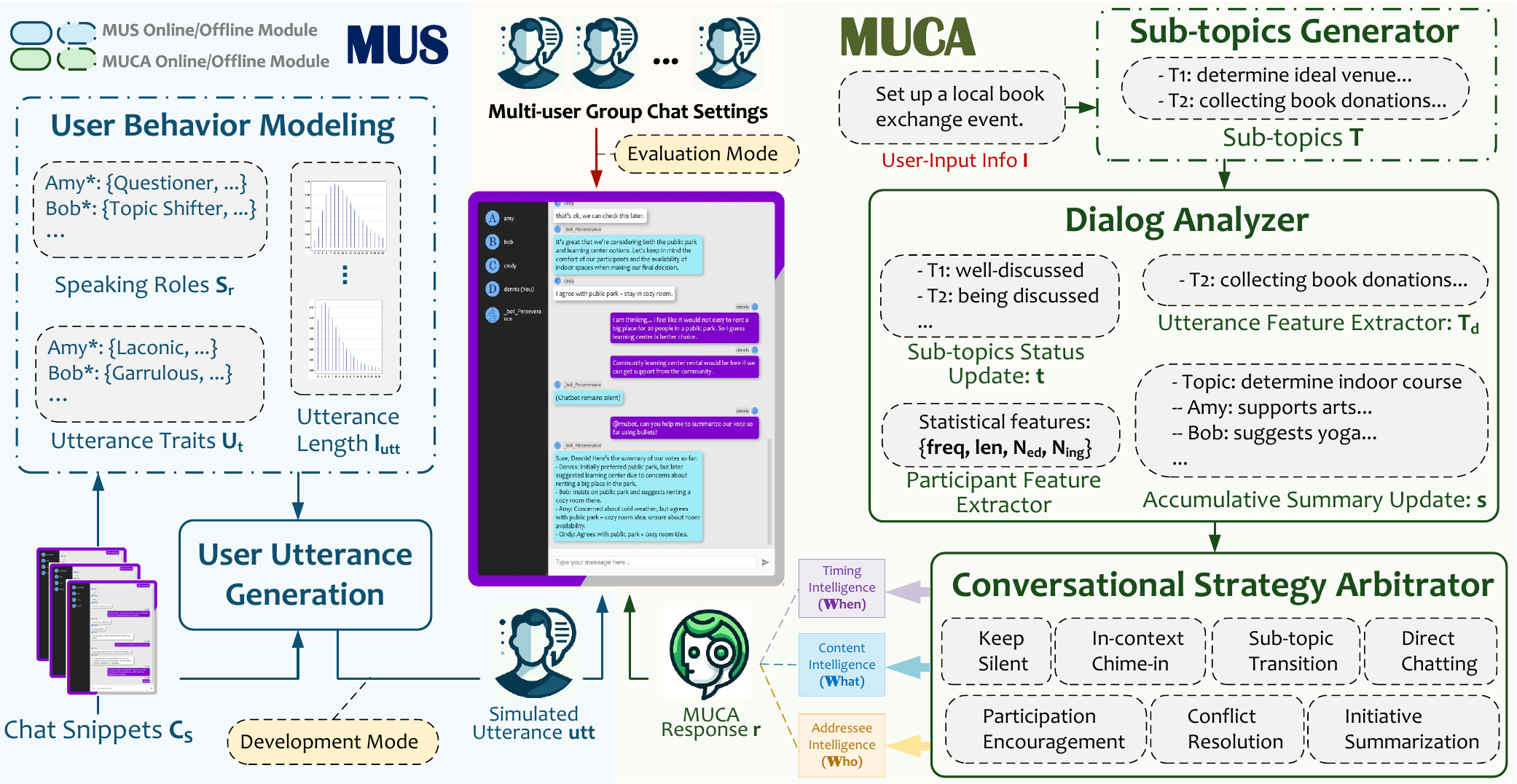}
\vspace{-0.5em}
\caption{Framework architecture, which is composed of the proposed MUCA (Sec. ~\ref{subsec:framework_architecture}) and MUS (Sec. ~\ref{subsec:user_simu}). The MUCA is periodically iterated via the interaction with the proposed MUS in the development mode, while real users are interacting with MUCA in the evaluation mode. The temporary results in the gray dash boxes serve as examples.}
\Description{A flowchart that shows the chat interface and multi-user chatbot background action logic.}
\label{fig:system_overview}
\end{figure*}

\begin{itemize}[left=6pt]

    \item We show that the proposed MUCA enhances the multi-user chat experience by controlling the \emph{3W (What, When, Who)} dimensions through its three key modules (Subtopic Generator, Dialog Analyzer, and Utterance Strategies Arbitrator), enabling cohesive conversations with deeper context awareness.
    
    \item We propose MUS, a user simulator designed to mimic real user behavior and simulate dialogues between multiple participants. MUS facilitates the optimization of MUCA by enabling agent interactions that incorporate the "human-in-the-loop" approach.
    
    \item We evaluate MUCA through case studies and user studies across various tasks and group sizes. The results show that MUCA consistently outperforms a baseline chatbot in tasks such as decision-making, problem-solving, and open discussions.
    
\end{itemize}

\section{Related Work}
\label{sec:related_work}
LLMs, such as GPTs \cite{openai2023gpt4, ouyang2022training, brown2020language}, have demonstrated superior performance on various tasks. Moreover, the development of LLMs has sparked interest in chatbot research and enabled various applications built around LLM-based chatbots.

{\bf Single-user Chatbots:}
There has been significant exploration of the pre-training or fine-tuning of LLMs for task-oriented dialogue systems. Studies such as \cite{budzianowski2019hello, hosseini2020simple, yang2021ubar, su2021multi, wang2022task} have employed LLMs, pre-trained or fine-tuned on dialogue data, to develop dialogue models or chatbots for various domains and tasks, such as travel tickets booking or restaurant reservation, etc. However, these work typically focus on single-user scenarios. 

{\bf Multi-user Chatbots:} 
Most research on multi-party or multi-user dialogue systems \cite{ouchi2016addressee, zhang2018addressee, gu2021mpc, gu2023gift, song2022supervised} have been focusing on training models on multi-party conversation datasets for the following tasks: addressee recognition, speaker identification, response selection and generation. Beyond these tasks, there are other important dimensions that have been explored when designing multi-party chatbots. For example, \cite{do2022should} proposed balanced participation communication strategies, and \cite{wagner2022comparing} presented four moderation strategies for planning and negotiating joint appointments. \cite{kim2020bot} described four features that can aid in facilitating group discussions.

Different from the above work, our MUCA handles the above tasks and design dimensions in a unified framework. The framework is based on LLMs, such as GPT-4, which has shown comparable performance in zero-shot settings to supervised models trained on multi-party datasets \cite{tan2023chatgpt}. In addition, MUCA relies on prompting methods \cite{wei2022chain, wang2022self, yao2023tree, wang2023planandsolve} to improve the capability of LLMs across various design dimensions, avoiding the need for fine-tuning and data collection. It can also be easily configured for different dialogue scenarios by updating the conversational strategy modules.

{\bf Multi-user Robots:} There has also been extensive research on multi-user human-robot interactions \cite{inoue2021multi, inoue2020attentive, skantze2021turn, sebo2020robots} based on acoustic and visual signals. However, these signals are generally not available in the text-based chatbots that our work focuses on.

\section{Framework Architecture}
\label{sec:methodology}

\subsection{Design Dimensions and Challenges}
\label{subsec:design_dimensions}
In this section, we describe the \emph{"3W"} dimensions for multi-user chatbots and the challenges MUCA addresses. While we believe \emph{"3W"} dimensions is applied broadly to varied group chats, the challenges presented can differ by scenario. This paper specifically focuses on chatbots that act as an assistant for multi-user conversations, similar to prior rule-based multi-user chatbots \cite{Cranshaw_2017, Avula2018SearchBotsUE, Toxtli_2018}.

\subsubsection{3W Design Dimensions:} Single-user chatbot scenarios often follow the "adjacency-pair" structure in which one utterance from the user anticipates a response from the chatbot \cite{Schegloff1968SequencingIC}. Therefore, the primary metric for evaluating single-user chatbots focuses on content, or the \emph{"What"} dimension. Designing chatbots for multiple users is far more challenging due to the \emph{3W (What, When, Who)} dimensions of the design space, which are the content, timing, and recipient of the response, as detailed below:

\begin{itemize}[left=6pt]

\item \textbf{Content Intelligence (\emph{"What"}):} It relates to what chatbots should respond, and can be more complex in multi-user cases due to the need to address challenges such as conflict resolution and multi-threaded discussions with multiple users.
\item \textbf{Timing Intelligence (\emph{"When"}):} It relates to whether chatbots are able to respond at the right timing or stay silent as needed.
\item \textbf{Addressee Intelligence (\emph{"Who"}):} It relates to whom the chatbots should respond, such as a specific group of participants, unspecified participants, or all participants.
\end{itemize}


\subsubsection{Design Challenges:} This paper focuses on five design challenges, which are linked with at least one of the \emph{"3W"} dimensions, depicted in Fig.~\ref{fig:challenges}. They are detailed below follows:

\begin{itemize}[left=6pt]

\item \textbf{Stuck Conversation Advancement:} MUCA can identify and appropriately chime in when a conversation is stuck, e.g., where the users were trying to estimate the transportation cost. It is closely related to the dimensions of \emph{"When"} and \emph{"What"}.
\item \textbf{Multi-threaded Discussion Management:} MUCA can handle concurrent topics and identify the participants involved in each topic, e.g., users are discussing cost estimation and venue selection at the same time in Fig.~\ref{fig:challenges}. It is related to the dimensions of \emph{"What"} and \emph{"Who"}.
\item \textbf{Responsiveness Requirement:} By carefully managing the chime-in rate, MUCA aims to provide reasonable responsiveness under the potentially high message traffic and complex interactions presented in multi-user chats. It is particularly related to \emph{"When"} dimension as the capability of responding in a timely manner is essential to perform time-sensitive tasks.
\item \textbf{Participation Evenness:} MUCA is intentionally designed to encourage even participation by identifying inactive users, e.g., Dennis in Fig.~\ref{fig:challenges}) and determining the proper timing for intervention and customized encouragement. It is relevant to all \emph{"3W"} dimensions.
\item \textbf{Conflict Resolution:} MUCA is capable of offering recommendations to assist participants in reaching an agreement during voting, resolving disputes (e.g., for Amy and Cindy in Fig.~\ref{fig:challenges}), or concluding discussions. It is related to all \emph{"3W"} dimensions.
\end{itemize}

The above challenges are neither required nor comprehensive but represent a design choice for this work. Participation Evenness and Conflict Resolution are common when conducting goal-oriented discussions in multi-user settings \cite{do2022should, kim2020bot}. Multi-threaded Discussion Management follows similar idea in \cite{Toxtli_2018} to track the status of tasks for each user. The rest of the two proposed design challenges are horizontal for the chat assistant in both chitchat and goal-oriented group chats. MUCA is a flexible design framework wherein targeted challenges can be adjusted by configuring the modules.

\subsection{Multi-User Chat Assistant (MUCA)}
\label{subsec:framework_architecture}

This paper defines several terms: $p_{\theta}$ as a pre-trained LLM with parameter $\theta$, $p_{\theta}^{CoT}$ as $p_{\theta}$ with Chain-of-Thoughts (CoT) integration \cite{wei2022chain}, $I$ and $T$ for user input and derived sub-topics, and $t$, $s$ for the status of $T$, and utterance summaries. Simplified notation $y \sim \bar{p}_{\theta}(y|v_1, v_2, \dots)$ indicates sampling and post-processing from LLM pdf $p_{\theta}$ for output $y$, given inputs $v_k$ to a prompt template. $U_{N, i}$ denotes $N$ most recent utterances at time $i$. Two context window sizes are used: short-term $U_{N_{sw}, i}$ and long-term $U_{N_{lw}, i}$, where $N_{lw}=10*N_{sw}$. $P$ represents the total number of users. 

The proposed MUCA consists of three major modules, depicted in Fig.~\ref{fig:system_overview}: (1) the Sub-topics Generator initializes the relevant sub-topics based on the user-inputs information before the chat starts; (2) the Dialog Analyzer then extract useful signals from the conversation, enabling MUCA to comprehend the ongoing conversation; and (3) the Conversational Strategies Arbitrator selects the appropriate strategy based on the signals from Dialog Analyzer and finally generate the response. Thus, they are sequentially executed when the chat begins. Three major modules are described below.

\subsubsection{Sub-topics Generator:}\label{subsubsec:subtopic_gen} This LLM-based module initiates relevant sub-topics $T$ (e.g., venue selection and book donations), based on the user-input information $I$ (e.g., set up a book exchange event): $T \sim \bar{p}_{\theta}(T|I)$, as shown in Fig.~\ref{fig:system_overview}. It enables the MUCA to smoothly engage in conversation based on derived sub-topics.


\subsubsection{Dialog Analyzer:} \label{subsubsec:dialog_ana} Its major task is to extract useful signals, assisting the Conversational Strategies Arbitrator in selecting a suitable conversational strategy for response. 

\begin{itemize}[left=6pt]
    \item \textbf{Sub-topic Status Update:} By using CoT style prompting, this sub-module categorizes the current status of each sub-topic $t$ as three statuses, namely, \emph{not discussed}, \emph{being discussed}, or \emph{well discussed}: $t_{i+1}, ts_{i+1}\sim \bar{p}_{\theta}^{CoT}(ts_{i+1}, t_{i+1} |I, t_{i}, ts_{i}, U_{N_{sw}, i})$, where topic summaries $ts$ is firstly generated to help track progress and enhance outcomes. 
    

    \item \textbf{Utterance Feature Extractor:} It extracts \emph{being discussed} sub-topics $T_{d}$ using context $U_{N_{sw}, i}$ from all sub-topics $T$: $T_{d} \sim \bar{p}_{\theta}(T_{d} | T, U_{N_{sw}, i})$ where $T_{d} \subset T$, (e.g., collecting book donations in Fig.~\ref{fig:system_overview}). It enables MUCA to track current sub-topics especially in the multi-threaded discussions mentioned in Sec.~\ref{subsec:design_dimensions}.

    \item \textbf{Accumulative Summary Update:} It updates the summary for each user across various sub-topics for full chat history,\footnote{Modern LLMs may process over 32k tokens, enabling LLM-based chatbots to use long historical data, despite efficiency and cost concerns. Our work uses a smaller context window $U_{N_{sw}, i}$ to accumulative update the summary: $s_{i+1} \sim \bar{p}_{\theta}(s_{i+1} | T_{d}, s_{i}, U_{N_{sw}, i})$, showing that summarization is feasible for LLMs with smaller windows.} which essentially builds a memory into the MUCA system.

    \item \textbf{Participant Feature Extractor:} It extracts statistical features like chime-in frequency $freq$, utterance total length $len$ per user from $U_{N_{sw}, i}$ and $U_{N_{lw}, i}$, which serves as a reference for customizing encouragement to increase lurkers' participation. The number of participants who discussed the sub-topic from the beginning $N_{ed}$ and the number of ongoing participants under the short-term context window $N_{ing}$ serve as signals for Sub-topic Transition in Conversational Strategies Arbitrator.
    
\end{itemize}

\subsubsection{Conversational Strategies Arbitrator:} \label{subsubsec:conv_strategy_arbitrator} As shown in Fig.~\ref{fig:system_overview}, MUCA interacts with users through seven pre-defined conversational strategies. Among them, Initiative Summarization and Sub-topic Transition are proven to be helpful in multi-user settings \cite{kim2020bot}. Besides, In-context Chime-in, Keep Silent and Direct Chatting are proposed to help address the challenges of Stuck Conversation Advancement and Responsiveness Requirement to maintain the chat flow.

Conversational strategies are ranked dynamically and their default ranking is presented below. The highest-ranked one is chosen among all eligible strategies whose trigger conditions are met. The response $r$ is generated using current summary $s$, the $U_{N_{sw}, i}$ and other upstream signals $sig$.

\begin{itemize}[left=6pt]

    \item \textbf{Direct Chatting:} It enables participants to directly interact with MUCA, which serves as a support assistant for individual users, addressing their specific requests as needed. It always has the highest priority and MUCA responds immediately regardless of the execution period once a user pings the MUCA. Many upstream features are extracted by the Dialog Analyzer and used as references for generating appropriate responses: $sig = \{t, T_{d}, I\}$. It is also worth mentioning that additional well-crafted prompting is required to avoid potential hallucination\footnote{When a chatbot is designed based on LLMs, hallucination issues may be inherited, generally causing confusion and misunderstanding for users. Without careful treatment, the chatbot might provide irrelevant or incorrect information.}, which is very common especially in this conversational strategy. Examples can be found in Sec.~\ref {subsec:case_study}.

    \item \textbf{Initiative Summarization:} It creates a take-home summary from chat messages, offering an insightful understanding of the discussion. Its trigger condition is heuristically designed for the scenarios when enough participants $N_{active}$ actively joined discussions since the last triggering. Accumulative Summary Update sub-module periodically updates the summary using $sig = \{T_{d}\}$ and concisely presents the key take-home message, which will be displayed when Initiative Summarization becomes the highest ranked eligible conversational strategy.

    \item \textbf{Participation Encouragement:} It aims to engage less vocal participants and promote balanced contributions in a conversation. The process of identifying a participant as a lurker is designed to be conservative. A participant is only considered as a lurker if their $freq$ and $len$ are significantly lower than the average in the long-term context window, and they have also spoken very little in the $U_{N_{sw}, i}$. Instead of using measures like KL divergence which evaluates overall distribution difference, we compute a ratio related to the variance to focus on individual participant data.

    \item \textbf{Sub-topic Transition:} It introduces a new, relevant topic when the current one is well-discussed or loses interest among most users. Note that its priority is lower than Participation Encouragement since MUCA encourages lurkers to contribute before considering to start a new sub-topic using $sig = \{N_{ed}, N_{ing}\}$. Introducing a new sub-topic may disrupt the conversation flow and potentially divert the discussion from its current status.

    \item \textbf{Conflict Resolution:} It helps users reach a consensus in a timely manner, thereby providing an efficient discussion procedure. Different from previous studies which set time limitations for each task \cite{kim2020bot}, MUCA provides suggestions to help parties with diverse opinions reach a consensus, and at the same time suggests a next topic for discussion, see example in Sec.~\ref{subsec:case_study}. Its trigger condition is met when the number of well-discussed sub-topics does not increase for a given period of time.

    \item \textbf{In-context Chime-in:} It offers an automatic chime-in mechanism to enhance conversation depth by providing insights, advancing stuck scenarios, and addressing users' concerns. Its trigger condition is controlled by two factors: (1) \emph{silence factor probability}: it increases with the number of consecutive silent turns; and (2) \emph{semantic factor probability}: it is associated with situations where the conversation is stuck due to repetitive utterances or unresolved issues that the chatbot must address. It uses the same $sig$ as Direct Chatting as it also needs to provide information that requires the long-term context.
    
    \item \textbf{Keep Silent:} It is automatically activated when other trigger conditions are not met, maintaining the conversation's flow without distracting participants.
    
\end{itemize}

\subsection{Multi-User Simulator (MUS)}
\label{subsec:user_simu}
In dialogue systems, chatbots can interact with users for training data collection \cite{Su16Online}, which can be costly and time-consuming. To expedite MUCA's training and development, we propose an LLM-based MUS that emulates user behavior, simulating dialogues for virtual users and facilitating optimization for MUCA, illustrated in Fig.~\ref{fig:system_overview}. Also, by incorporating a "human-in-the-loop" approach, MUS uses human feedback to refine its own prompts, thereby enhancing simulation outcomes. MUS comprises two main modules:

\textbf{User Behavior Modeling:} It processes $C_{s}$, chat snippets derived from real chat records to obtain: speaking role $S_{r}$, utterance traits $U_{t}$ and utterance length $l_{utt}$. It executes once before the simulation.

\textbf{User Utterance Generation:} This module uses $S_{r}$, $U_{t}$, $l_{utt}$, and signals in context window to produce natural language utterances $utt$, which mimics real user behavior from the chat snippets $C_{s}$.

\begin{figure*}[!ht]
\vspace{-0.2em}
\centering
\includegraphics[width=1.0\textwidth]{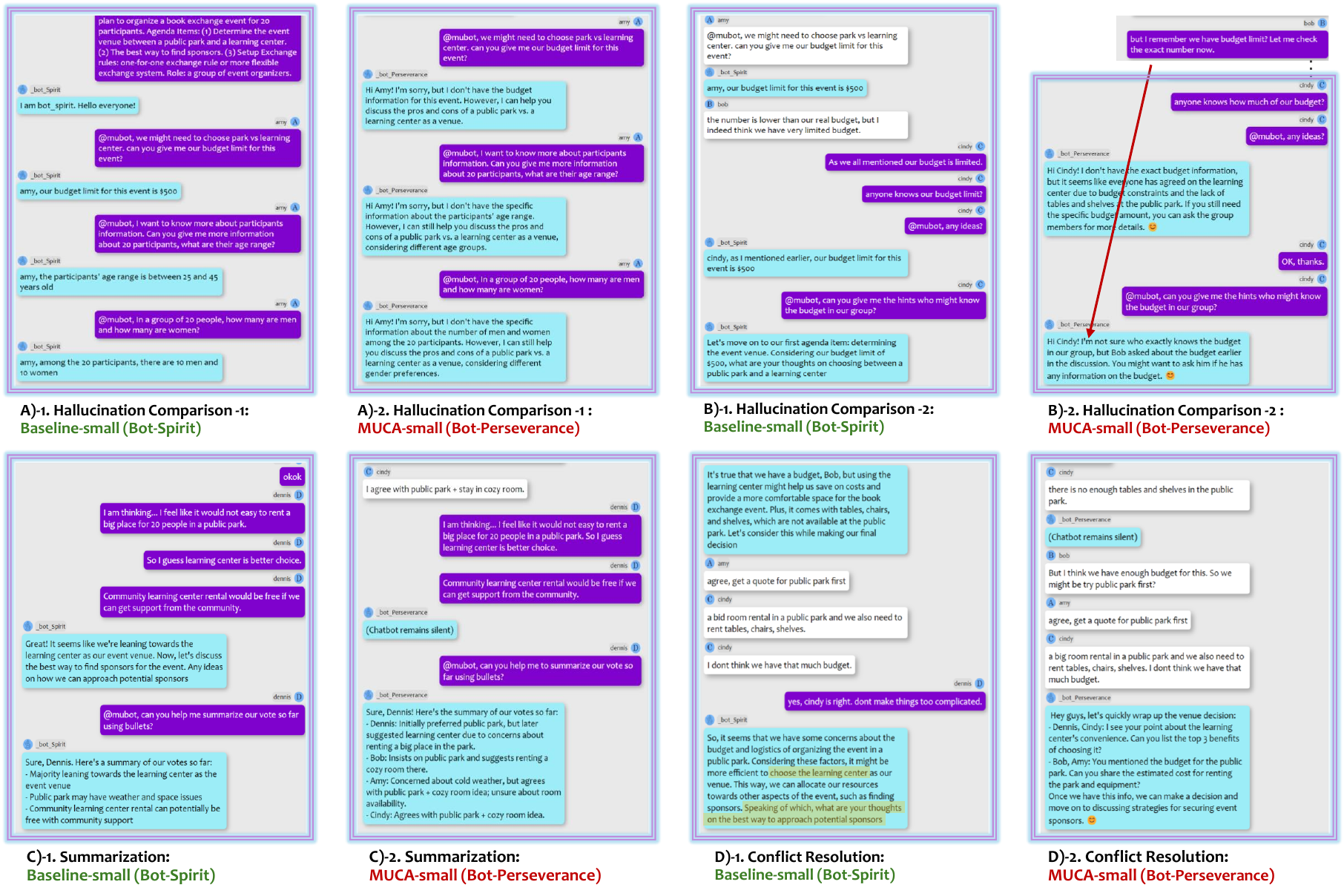}
\vspace{-2em}
\caption{Qualitative comparison between \emph{Baseline-small} and \emph{MUCA-small}: A), B) hallucination issues, C) summarization feature, and D) conflict resolution capability. The conversation consists of 1 chatbot (\emph{\_bot\_Spirit} for \emph{Baseline-small} or \emph{\_bot\_Perseverance} for \emph{MUCA-small}) and 4 participants, namely, Amy, Bob, Cindy, and Dennis. }
\Description{Eight chat snippets that illustrate how two models handle four representative problems.}
\label{fig:case2}
\end{figure*}

\section{Evaluation}
\label{sec:evaluation}
We built a group chat system with the support of multi-user chatbot and conducted case and user studies to evaluate MUCA's performance across different topics and group sizes.

\subsection{Experimental Configuration}

\label{subsubsec:chatbot_config}
This section evaluates a baseline model based on GPT-4~\cite{openai2023gpt4} and our proposed MUCAs with slightly different configurations for various group sizes. A general description of the baseline system and two proposed MUCAs~\footnote{In this section, aliases \emph{\_bot\_Spirit}, \emph{\_bot\_Perseverance}, and \emph{\_bot\_Discovery} were given to \emph{Baseline-small}, \emph{MUCA-small}, and \emph{MUCA-medium} in user studies, respectively. It ensures that participants in user studies do not have prior knowledge of each chatbot, thereby preventing biases.} are as follows:

\textbf{Baseline-small:} GPT-4 with a single prompt, which takes user-input information, conversation context, and users’ names as input and outputs generated responses. In the prompt, we simply define its conversational strategies, for example, \emph{keep silent}, \emph{direct chatting}, and \emph{in-context chime-in}. This version is applied in a 4-person group chat with a short-term context window size ($N_{sw}$) of 8 and an execution interval~\footnote{To ensure performance-efficiency and amid the high message traffic in complex multi-user interactions, Dialog Analyzer and Conversational Strategies Arbitrator are sequentially executed for every $N_{exe}$ utterance.} ($N_{exe}$) of 3.

\textbf{MUCA-small:} GPT-4 with the MUCA framework. It is applied in a 4-person group chat, and uses the same configuration ($N_{sw}, N_{exe}$) and user-input information as \emph{Baseline-small}.

\textbf{MUCA-medium:} It shares the same framework and architecture as \emph{MUCA-small} but has different configurations. These configurations are automatically adjusted based on the number of participants ($N_{sw} = 2*P$, $N_{exe}= 0.75*P$) to maintain the latency-efficiency for an 8-person group chat.

For evaluation, we focus on 4 goal-oriented communication tasks (i.e., estimation, decision-making, problem-solving, and open discussion) rather than chit-chat. Specifically, we designed four discussion topics, where Topic-A (\emph{"indoor course set up in a community learning center"}) and Topic-B (\emph{“interview agenda for hiring arts instructors”}) are used in the user studies (in Sec.~\ref{subsec:user_study}), while Topic-C (\emph{"organize book exchange event"}) and Topic-D (\emph{"organize a hiking activity for 50 members"}) are used in the case study (in Sec.~\ref{subsec:case_study}).

These topics require users to complete the tasks collaboratively and reach agreements, and MUCA is anticipated to aid participants in fostering comprehensive thinking and improving chat efficiency.

\subsection{Case Study}
\label{subsec:case_study}

We qualitatively show \emph{MUCA-small}'s key functions using case studies. We focus on comparing \emph{MUCA-small} against \emph{Baseline-small} in handling factuality  
 hallucination, summarization, and conflict resolution, as shown in Fig.~\ref{fig:case2}.
\begin{itemize}[left=6pt]

    \item \textbf{Factuality  Hallucination}: As shown in Fig.~\ref{fig:case2}-A)-1 and A)-2, \emph{Baseline-small} fabricated the information beyond user inputs (topics, hints, and agenda) and conversation history, such as budget limit, participants' age, and genders, potentially leading to distrust and bias. On the contrary, \emph{MUCA-small} flagged out-of-scope questions and aligned its responses with user inputs.

    We dive deeper into this issue in Fig.~\ref{fig:case2}-B)-1 and B)-2. For the unknown budget information, \emph{Baseline-small} fabricated a budget number, which Bob later corrected. Despite this, when Cindy inquired further, it stuck to the false info and even attempted a topic shift. In contrast, \emph{MUCA-small} correctly inferred that Bob likely knew the budget based on his prior input. This highlights the complexity of processing multi-user chat history, relationships, and interactions, which pose challenges for generating accurate, hallucination-free responses. Addressing these issues requires careful prompting design, even with a powerful LLM.
    
    \item \textbf{Summarization}: As shown in Fig.~\ref{fig:case2}-C)-1 and C)-2, \emph{Baseline-small} failed to understand the query intent from Dennis, which was summarizing the votes from all participants. Instead, it summarized opinions, and its summary was inaccurate due to the limited context window by design. For example, it mentioned the "Majority" leaning towards the learning center, but actually only Dennis voted for this option. In contrast, \emph{MUCA-small} overcame window size limitations, and correctly summarized and categorized votes by users.

    \item \textbf{Conflict Resolution}: In multi-user chatting environment, diverse opinions are common. As shown in Fig.\ref{fig:case2}-D)-1, \emph{Baseline-small} attempted to resolve conflicts with its own biased opinion and even attempted shifted topics, disrupting the conversation flow. In contrast, Fig.\ref{fig:case2}-D)-2 shows \emph{MUCA-small} summarizing differing views, raising thought-provoking questions, and resolving conflicts where possible.

\end{itemize}

\begin{figure*}[!ht]
\centering
\includegraphics[width=0.925\textwidth]{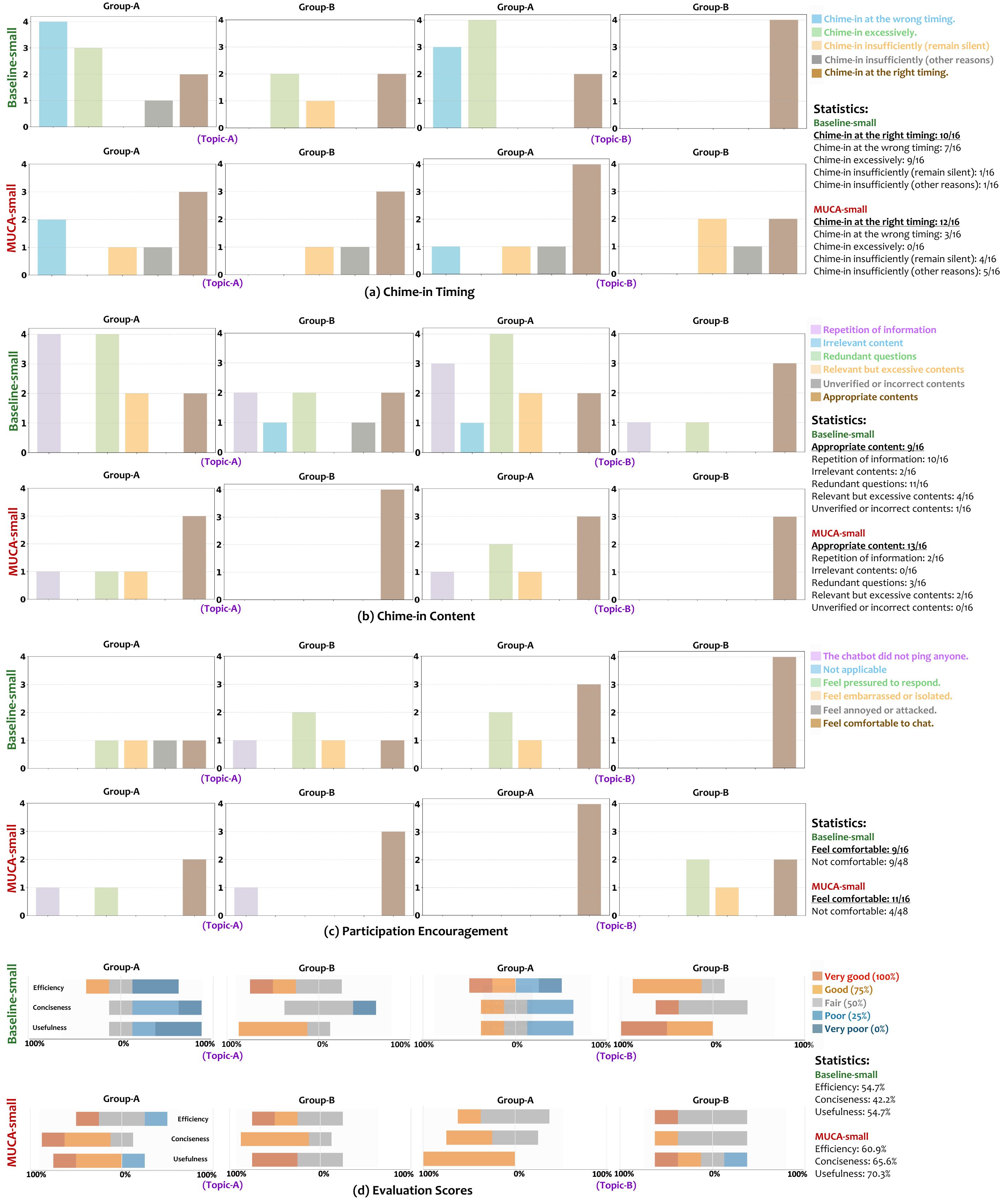}
\vspace{-1em}
\caption{A comparison between \emph{Baseline-small} and \emph{MUCA-small}. Each set of results presents the performance of \emph{Baseline-small} and \emph{MUCA-small} in two separate rows. In (a)-(c), each bar chart illustrates the counts of options selected by users if they ever encountered these scenarios during the chat. The accompanying statistics on the right-hand side summarize the counts in each row. In (d), users rated each chatbot on efficiency, conciseness, and usefulness, using options from "Very Good" to "Very Poor". Corresponding scores are displayed on the right.}
\Description{Charts that show the comparison results of two models.}
\label{fig3:comparison}
\end{figure*}

\subsection{User Study}
\label{subsec:user_study}

We conducted user studies to qualitatively and quantitatively compare the effectiveness of \emph{MUCA} against \emph{Baseline-small}.

\subsubsection{Study Design and Procedure:} We conducted user studies with three participant groups, two small groups (Group-A and Group-B with 4 people each) and one medium group (Group-C with 8 people), maintaining a 1:1 ratio of females to males. As mentioned in Sec.~\ref{subsubsec:chatbot_config}, we chose two goal-oriented topics. The small group experiments compared \emph{Baseline-small} with \emph{MUCA-small}, while the medium group experiment demonstrated the MUCA's capabilities in more complex chatting scenarios in a larger conversation group.

In small-group experiments, Group-A tested Topic A first with \emph{Baseline-small} then \emph{MUCA-small}, and Topic B in the reverse order. Group-B reversed the chatbot order in the experiments to counter the learning effect, where participants might become more familiar with the topic after interacting with the first chatbot. Additionally, MUCA was applied to a medium group (Group-C) using Topic-A, demonstrating its effectiveness in the larger conversation group.

\begin{table*}
\caption{Comparisons in terms of quantitative results (upper three rows) and evaluation scores (bottom four rows).}
\centering
{
\renewcommand{\arraystretch}{1.2}
\begin{tabular}{c|ccc|cc}
\Xhline{2\arrayrulewidth}
         \multirow{2}{*}{\textbf{Metrics}}  & \multicolumn{3}{c|}{\textbf{Indoor Course (Topic-A)}} & \multicolumn{2}{c}{\textbf{Interview Agenda (Topic-B})} \\
                            \cline{2-6}
                           & \emph{Baseline-small} & \emph{MUCA-small} & \emph{MUCA-medium} & \emph{Baseline-small}     & \emph{MUCA-small}     \\
                            \hline
Engt.-Words/Conv.  &  426.5         &  531.5        &  875          &   531          & 636.5           \\
Engt.-Words/Utt.   &  7.23          &  8.93         &  8.75         &   8.85         & 11.27           \\
Evenness         &  106.6 $\pm$ 67.6\%  &  132.9 $\pm$ 47.1\% &  109.4 $\pm$ 56.0\% &   132.8 $\pm$ 58.0\% & 159.1 $\pm$ 61.2\%    \\
Consensus (\%)   &  50            &  66.7         &  66.7         &   50           & 100             \\
\Xhline{2\arrayrulewidth}

Efficiency  (\%)                &  50     &   62.5   &    68.75  &    59.38   & 59.38            \\
Conciseness (\%)                &  31.25  &   71.88  &    75     &    53.13   & 59.38            \\
Usefulness (\%)                 &  43.75  &   71.88  &    65.63  &    65.63   & 68.75            \\
Overall Rating (\%)             &  37.5   &   69.44  &    69.44  &    52.78   & 63.89            \\

\Xhline{2\arrayrulewidth}
\end{tabular}
}

\label{tab:compare}
\vspace{-1em}
\end{table*}

\subsubsection{Comparison in Small-size Groups}
\quad

\textbf{Statistics from Users:} Fig.~\ref{fig3:comparison} presents a quantitative comparison of \emph{MUCA-small} and the baseline \emph{Baseline-small} across four aspects:


\textbf{Chime-in Timing:} Both chatbots have ever chimed in at the good timing at least once during the whole conversation, while \emph{MUCA-small} performs slightly better, as demonstrated in Fig.~\ref{fig3:comparison}. Notably, 56.25\% (9 out of 16) participants felt that \emph{Baseline-small} chimes in excessively. This is believed to be a result of its less strategically designed behavior -- it always replies every three turns ($N_{exe}=3$) and ignores the "keeping silent" instruction in its prompt, as described in Sec.~\ref{subsubsec:chatbot_config}. In contrast, such excessive chiming in was not reported for \emph{MUCA-small}. However, some participants noted that \emph{MUCA-small} occasionally chimed in infrequently, constrained by $N_{exe}$ and "keeping silent" policy. Adjusting $N_{exe}$ poses a common design trade-off between latency and user experience.


 \textbf{Chime-in Content:} \emph{MUCA-small} generally offers appropriate responses (13 out of 16) with infrequent inappropriate content, as shown in Fig.~\ref{fig3:comparison}. In contrast, \emph{Baseline-small} often repeats the information, asks redundant questions, and generates excessive content. While some information might be useful, it can overwhelm participants, requiring extra effort to discern valuable content.
 

 \textbf{Participation Encouragement:} The interaction feature, i.e., pinging a lurker by a chatbot, should be cautiously designed, including its chime-in timing, frequency, and contents. It may impose negative feelings on participants, while a good design may improve user engagement. As shown in Fig.~\ref{fig3:comparison}, \emph{MUCA-small} has a better user experience in terms of comfortableness over \emph{Baseline-small}.

 \textbf{Evaluation Scores:} Three additional metrics are applied in user studies, as shown in Fig.~\ref{fig3:comparison}. \textbf{Efficiency} refers to the chatbot's timely responses; \textbf{Conciseness} refers to the chatbot's on-point and non-redundant response; \textbf{Usefulness} refers to whether its responses are helpful or insightful. \emph{MUCA-small} achieved significantly higher ratings in these user-friendly factors.



\subsubsection{Quantitative Study in Small-size Groups}
\label{subsec:quant_small_group}
The quantitative comparisons for two chatbots are shown in Table \ref{tab:compare}. User engagement (abbreviated as Engt.) is compared with two metrics, the average number of words exchanged per conversation (Engt.-Words/Conv.) and the average number of words per utterance (Engt.-Words/Utt.). Evenness is assessed by calculating the sample standard deviation (STD) of the word count input by each participant, expressed as a percentage of the mean. The consensus is obtained from the rates given by Group-A and Group-B for small-size experiment or Group-C for medium-size experiment, where the rate is represented by the number of agreements reached over the total number of tasks.

From the comparison in Table~\ref{tab:compare}, \emph{MUCA-small} helps participants get better engagement, shown by increased Engt.-Words/Conv. and Engt.-Words/Utt., which indicates that participants were more inclined to engage in extensive conversations and to compose longer utterances. \emph{MUCA-small} enhances evenness in Topic A discussions with a lower STD while keeping similar evenness in Topic B with a comparable STD over \emph{Baseline-small}. \emph{MUCA-small} achieves a higher consensus rate than \emph{Baseline-small} thanks to its less frequent interruptions maintaining efficient conversation flow, provision of practical suggestions aiding reaching agreement, and insightful comments that enhance efficient discussion. Conversely, \emph{Baseline-small} often revisits well-discussed topics and provides redundant information, resulting in inefficient discussion.

Additionally, Table~\ref{tab:compare} shows average scores from Group-A and Group-B on Efficiency, Conciseness, and Usefulness for small-size experiments and scores from Group-C for medium-size experiment. For Topic-A, \emph{MUCA-small} outperforms \emph{Baseline-small} with 12.5\%, 40.6\%, and 28.1\% higher scores on Efficiency, Conciseness, and Usefulness, respectively. \emph{MUCA-small} scores slightly higher in Topic-B. The Overall Rating also reflects similar trends: \emph{MUCA-small} surpasses \emph{Baseline-small} by 31.9\% in Topic-A and 11.1\% in Topic-B.

\subsubsection{Quantitative Study in Small-size and Medium-size Groups}
\label{subsec:quant_medium_small_group}
Managing conversations in medium-sized groups is more challenging than in small groups. A facilitator chatbot should be more effective in medium-sized groups, as it promotes even contribution among participants, countering social loafing and free-riding behaviors, which are common in larger groups. However, this increased participation raises the chatbot's cognitive load for organizing diverse opinions, making larger group management more complex.

We conducted a user study for a medium group and recorded its statistics in Table~\ref{tab:compare}. We find that \emph{MUCA-medium} maintains stable performance despite larger group sizes compared to \emph{MUCA-small}. Notably, increased Engt.-Words/Conv infers that larger groups yield more opinions. There is a subtle change in Engt.-Word/Utt due to unchanged user chatting habit. Compared to \emph{MUCA-small}, \emph{MUCA-medium} with higher STD has lower evenness due to a natural outcome of larger group dynamics. Medium group reaches the same consensus rate as small groups. These findings underscore \emph{MUCA}'s consistent performance across varied group sizes.

As shown in Table 1, participants in small and medium groups gave comparable user evaluation scores, while MUCA consistently outperforming \emph{Baseline-small}. The statistic results highlight \emph{MUCA-medium}’s effectiveness in managing larger group interactions.

\section{Conclusion}
\label{sec:conclusion}

In this work, we discussed the crucial \emph{3W} design dimensions, namely \emph{"What"} to say, \emph{"When"} to respond, and \emph{"Who"} to answer, for multi-user chatbot design. We identified challenges that are commonly faced in various chat scenarios. An LLM-based multi-user chatbot framework called MUCA was proposed to address these challenges. The paper also devised an LLM-based user simulator, named MUS, to speed up the development process for MUCA. Experimental results obtained from both case studies and user studies demonstrate the effectiveness of MUCA in goal-oriented conversations with a small to medium number of participants.
\section*{Limitations}
\label{sec:limitations}

LLMs do see many challenges, including those having significant societal implications such as bias, fairness, toxicity, etc., and we refer readers to the numerous studies that are dedicated to addressing these pressing problems. We emphasize that the present version of MUCA still faces many challenges around these issues with societal implications. For example, for users who prefer to stay quiet, MUCA’s pinging these users may bring stress or other negative feelings for them. Also, as another example, MUCA’s intervention to address harmful or detrimental chats remains very limited. We would like to welcome researchers to continue investing efforts on improving multi-user chatbots along these dimensions. For the remainder of this section, we will discuss other issues that are particularly relevant to MUCA and MUS.

\textbf{Multi-user Chat Assistant (MUCA):} The proposed MUCA is a pioneering work dedicated to multi-user chats. Although it is by no means a comprehensive solution, it provides significant insights that could pave the way for future work in this field. We have identified several challenges that call for further research:

 \begin{itemize}[left=6pt]
     \item Firstly, MUCA encompasses seven sub-modules dedicated to conversational strategies, but only the top-ranked one is chosen at a time for generating a response.  This approach overlooks the potential to validate the response's quality, as it is delivered irrespective of its merit. We believe that by requesting all the conversational strategy sub-modules to generate a response concurrently, MUCA will be able to comprehensively evaluate and validate all the response candidates. The final augmented response could then be synthesized by either selecting or merging from this pool of response candidates through another post-conversational-strategy procedure.
     \item Secondly, in our user study cases, we adjusted the hyper-parameters ($N_{exe}$, $N_{sw}$, $N_{lw}$, $W$, $C$, $f$ and $g$) in MUCA based on experimental  results on small to medium groups. For larger conversation groups, the effectiveness of the selected hyper-parameters needs empirically validation. Also, an automated mechanism determining these parameters based on the configurations and the environmental variables of the conversations can also greatly alleviate the burden of tuning these parameters. 
     \item Thirdly, compute resources requested by LLMs inference pose a significant constraint for MUCA, especially for large chat groups. To mitigate this challenge, we have slightly increased the execution interval ($N_{exe}$), which occasionally results in MUCA missing optimal opportunities for user interaction at the most suitable moment. Moreover, we have sometimes observed an interesting phenomenon wherein multiple participants simultaneously express the desire to directly engage with MUCA, leading to a surge in computational demands. How to handle high volume of LLM calls with limited compute resources, while simultaneously striving to preserve the responsiveness of MUCA to the best extend, is a topic that worth further investigation.
 \end{itemize}

\textbf{Multi-user Simulator (MUS):} Constructing a high-quality and specialized user simulator for a specific task can be a labor-intensive process \cite{walker1997paradise, liu2017iterative}. Similar to previous research, we also discovered that modeling human behavior is challenging for the user simulator: 

  \begin{itemize}[left=6pt]
     \item Firstly, generating natural language utterances with an LLM-based user simulator is challenging when utterances are short. For instance, the minimum length of utterance ($l_{min}=1$) and maximum length of utterance ($l_{max}=10$) extracted from chat history are quite small. To address this, we boosted $l_{min}$, $l_{avg}$, and $l_{max}$ for each virtual user correspondingly and also adjusted the number of words for the role of \emph{questioner}.
     \item Secondly, LLMs may not consistently follow instructions to generate a valid virtual user ID for the next turn to speak. Instead, it tends to predict the LLM agent to speak next, particularly when someone directly mentioned the LLM agent in the previous turn. To mitigate this issue, we randomly select the virtual user and their corresponding speaking role.
     \item Thirdly, virtual users suffer from repeating the same conversational strategy (e.g. asking questions, direct chatting) for consecutive turns. This issue might be due to the nature of the generative model which focuses on predicting the next token. To address this issue, we introduce a cool-down mechanism for some conversational strategies such as asking questions, direct chatting, and topic transition.
 \end{itemize}


\bibliographystyle{ACM-Reference-Format}
\bibliography{sample-base}

\appendix

\section{Appendix}
\label{sec:appendix}

\subsection{Prompting Example}
\label{subsec:app_module_detail}
\begin{figure*}[!ht]
\centering
\includegraphics[width=0.99\textwidth]{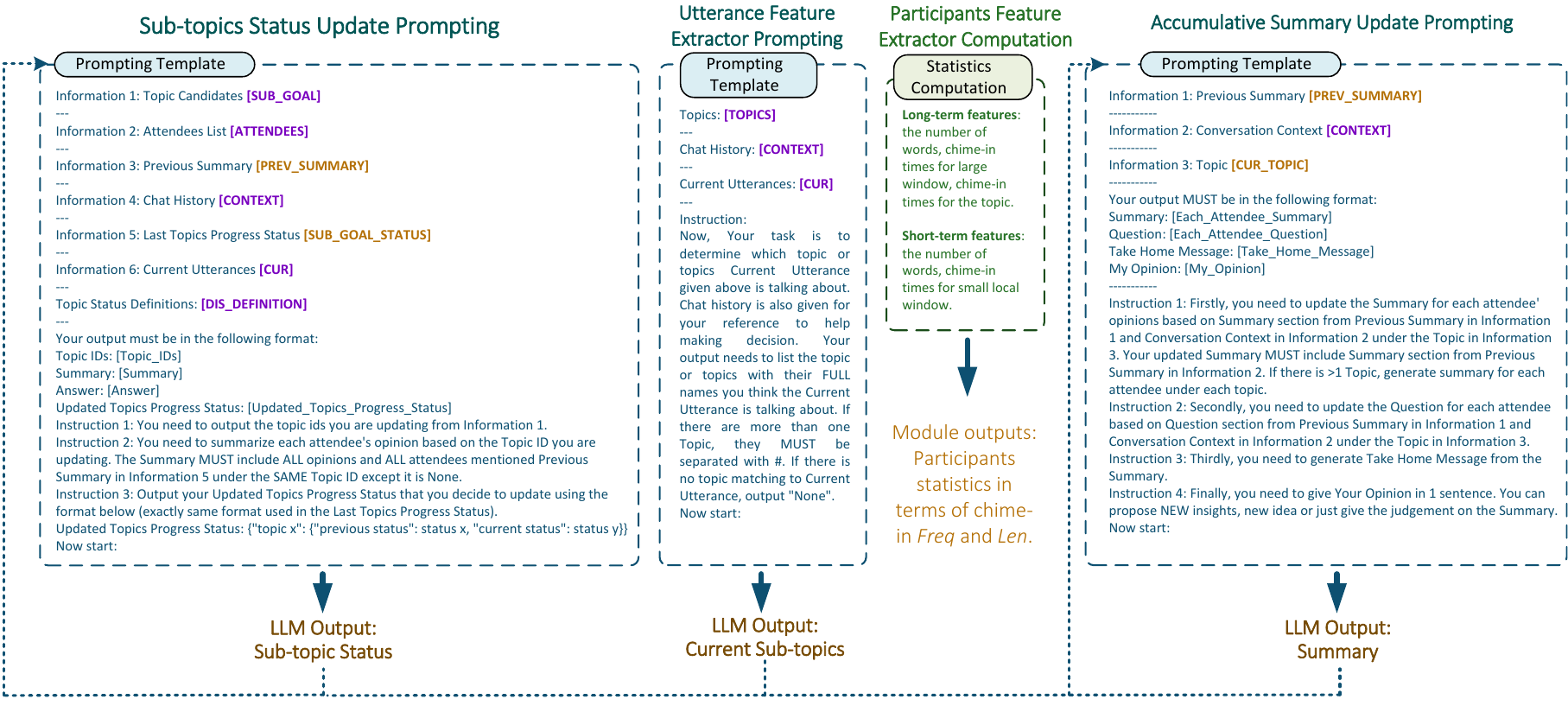}
\vspace{-1em}
\caption{Data flow for Dialog Analyzer, which includes \emph{participant feature extractor} and three LLM-based modules -- \emph{sub-topic status update, utterance feature extractor, and accumulative summary update}.  The placeholders (printed in purple) in the prompts are filled by sub-topics from the Sub-topic Generator, conversation signals such as attendee names, and utterances in $U_{N_{sw}, i}$. The generated outputs (sub-topic status, accumulative summary, and sub-topic \emph{being discussed}, all printed in yellow) will be fed back to the \emph{sub-topics status update} and \emph{accumulative summary update} as inputs for the execution in the next round.}
\label{fig:dialog_analyzer}
\end{figure*}

Fig.~\ref{fig:dialog_analyzer} shows the data flow for the Dialog Analyzer. Only the \emph{participants feature extractor} sub-module is based on statistical computation and the rest of the three sub-modules (\emph{sub-topic status update, utterance feature extractor}, and \emph{accumulative summary update}) are based on LLM inference results. Complete input prompt templates for the three LLM-based sub-modules where the purple and yellow text represent placeholders are shown. The purple ones are replaced by sub-topics from the sub-topic generator, conversation signals such as attendee names and utterances in the current context window, and the yellow ones are replaced by the generated outputs (sub-topic status, summary, and current sub-topic) from other modules. The outputs of the Dialog Analyzer will be fed into the downstream Conversational Strategies Arbitrator module to select the suitable conversational strategy for the response generation.

\subsection{System Design and Implementation}
\label{subsec:app_system_design}
The user interface (UI), designed with JavaScript, HTML, and CSS, is a static single-page web application that is responsible for managing user login and facilitating communication with the backend server. Upon initial access, the UI presents a login window and only denies entry if the username already exists. Additionally, the interface transmits user information and messages to the backend server while also broadcasting MUCA's messages received from the backend server, ensuring they are visible to all participants.

The backend server operates on a locally hosted machine for experimentation purposes. It leverages WebSocket protocol for bi-directional communications to enable multi-user conversations. The backend server is responsible for monitoring incoming messages from all users, and distributing these messages out to the other users. It also manages broadcasting system messages and processes login requests. The backend server maintains a record of all connected users, including MUCA, which is a special user that also communicates with other users via the backend.

MUCA establishes an asynchronous connection with the backend server. Upon receiving incoming messages from users, MUCA determines appropriate conversational strategies, taking chat history and relevant factors into consideration (as elaborated in Sec.~\ref{sec:methodology}). Subsequently, MUCA's response is sent back to the backend server for broadcasting. Incoming messages are accumulated in a queue and are processed periodically to better handle API call burst.

\begin{figure}[t]
\centering
\includegraphics[width=0.48\textwidth]{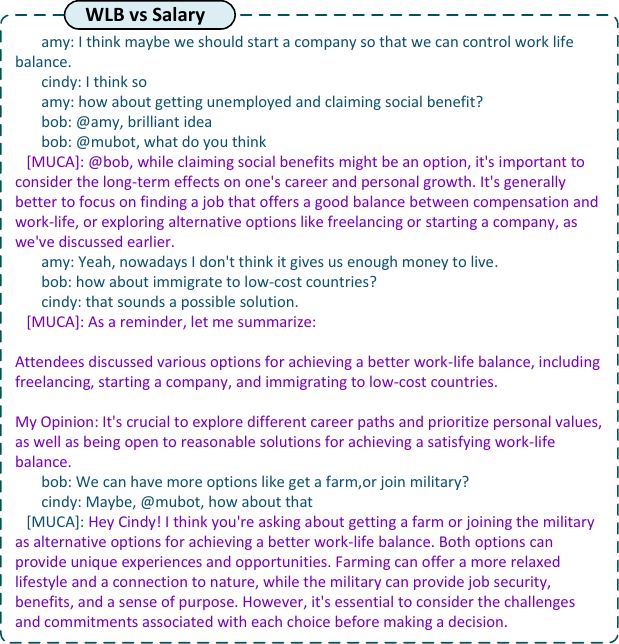}
\vspace{-1.5em}
\caption{An example of a group chat about a chit-chat topic on compensation and work-life-balance.}
\label{fig:wlb_salary}
\end{figure}

\subsection{Group Chat using Chit-chat Topic}
\label{subsec:app_chit_chat_results}
Fig.~\ref{fig:wlb_salary} shows the example for a chit-chat topic: 
\begin{itemize}[left=6pt]
    \item \textbf{Topic:} During today's chat, each attendee is going to share their personal choices of prioritizing compensation and work-life-balance when seeking a new job. Hint: encourage people to share their personal choices. Participant roles: they are seeking new jobs.
\end{itemize}

For chit-chat topics (non-goal-oriented communication), we found that MUCA does not play the same important roles as in goal-oriented communications, since the goal in chit-chat is sharing opinions rather than reaching agreements. In this context, summarizing, voting, or similar functionalities are less critical compared to goal-oriented conversations.

\subsection{Dialog Topics}
\label{subsec:app_dialog_topics}
We created four discussion topics, where Topic-A and Topic-B are used in the user studies (in Section~\ref{subsec:user_study}) and Topic-C and Topic-D are utilized in the case study (in Section \ref{subsec:case_study}). 
\begin{itemize}[left=6pt]
    \item \textbf{Topic-A:} During today’s chat, a group of attendees are going to set up a new indoor course in a community learning center for 20 college students. There are several sub-topics going to be discussed: (1) Determine the indoor course between arts, bakery, and yoga. (2) Set up a course format: a short, intensive course vs. a longer, more spread-out course. (3) Estimate the total costs for lecturers, given hourly pay ranges from \$16 to \$24 per lecturer. Participant roles: they are offering a new course in a community learning center.
    \item \textbf{Topic-B:} During today’s chat, a group of interviewers are going to set up a hiring interview composed of 2 sessions for a position of arts instructor for a senior community education program. There are several sub-topics going to be discussed: (1) Determine the format of 2 sessions, which can include traditional QnA, presentation, and resume scanning. (2) Determine the qualifying requirements: teaching experience vs. artistic accomplishments. (3) How to fairly take both recommendation letters and candidates' performance during the interview into the hiring decision process. Participant roles: they are going to interview arts instructors for senior community education.
    \item \textbf{Topic-C:} During today's chat, a group of event organizers are going to discuss the plan to organize a book exchange event for 20 participants. Agenda Items: (1) Determine the event venue between a public park and a learning center. (2) The best way to find sponsors. (3) Setup Exchange rules: one-for-one exchange rule or more flexible exchange system. Participant roles: they are event organizers.
    \item \textbf{Topic-D:} During today’s chat, a group of activity organizers are going to discuss the plan to organize a hiking activity in a mountain (3-hour driving) for 50 members (ages between 21-40) in a local hiking club. There are several sub-topics going to be discussed: (1) Estimate cost of transportation. (2) Find the best way to organize group sizes hiking start times, and locations to prevent congestion, considering the narrow portions of some trails. (3) The choices for trail difficulty – easy, medium, and hard. Participant roles: they are hiking activity organizers in the club.
\end{itemize}

\subsection{Future Work}
\label{appx:future_work}
The framework we propose for multi-user chatbots is not intended as a comprehensive solution for multi-user conversations. Rather, we hope this work can shed light on potential directions for future research in the field of multi-user chatbots. Several areas, including but not limited to the following, deserve further research:

\textbf{Component Orchestration:} MUCA integrates several components, enabling actions such as "participation encouragement" and "initiative summarization". These components have been carefully designed, tuned, and ranked to provide a harmonious experience to the chat participants. It can be beneficial to explore an easy plug-and-play method for users to design and incorporate new components into the framework without intensive tuning. Such a feature could be important, as different conversation scenarios may require chatbots to provide different set of functionalities.

\textbf{Human-in-the-loop Feedback Iteration:} Full user studies for feedback are costly and time-consuming. To continuously improve the chatbot post-launch, it is useful to collect implicit and explicit user behavior signals. This data should be easily transformable for automatic or semi-automatic chatbot enhancements.

\textbf{Rapidly Advancing AI Technologies:} The proposed MUCA framework is based on recent state-of-the-art LLMs, each with its unique style and best practices for prompting. It would be beneficial to investigate methods for updating the underlying AI models without the need of completely redoing prompting or component orchestration.

\textbf{Multi-modal Capabilities and External Resources:} As LLMs become increasingly capable of processing multi-modal data, a chatbot that interacts with multiple users using not only text, but also video, audio, and images is becoming feasible. Additionally, external resources could be integrated as a component for the chatbot to leverage to enhance the multi-user conversation experience.

\textbf{Multi-Chatbot Design:} The study concentrates on multi-user and single-chatbot interactions. However, scenarios involving interactions among multiple users and and multiple chatbots with different characteristics can merit further investigation.
For instance, in cross-disciplinary meetings, chatbots could serve as hosts, minute-takers, or subject matter experts, offering insights to human participants as needed.

\end{document}